\documentclass[letterpaper]{article} 
\usepackage{aaai24}  
\usepackage{times}  
\usepackage{helvet}  
\usepackage{courier}  
\usepackage[hyphens]{url}  
\usepackage{graphicx} 
\usepackage{natbib}  
\usepackage{caption} 
\usepackage[textsize=tiny]{todonotes}

\newcommand{\eg}{\emph{e.g. }}
\newcommand{\ie}{\emph{i.e.}}
\usepackage{amsmath}
\usepackage{amsmath}
\usepackage{amssymb}
\usepackage{mathtools}
\usepackage{amsthm}
\usepackage{tabularx}
\usepackage{bm}
\usepackage{tikz}
\usepackage{listings}
\usepackage{subcaption}
\usepackage{graphicx} 
\usepackage{algorithm}
\usepackage{bbding}
\usepackage{algorithmic}

\usepackage{fontawesome5}
\usepackage{newfloat}
\usepackage{listings}
\DeclareCaptionStyle{ruled}{labelfont=normalfont,labelsep=colon,strut=off} 
\floatstyle{ruled}
\newfloat{listing}{tb}{lst}{}
\floatname{listing}{Listing}
\title{Data Adaptive Traceback for Vision-Language Foundation
Models in Image Classification}
\author{
    Anonymous submission
}
\author {
    Wenshuo Peng\textsuperscript{\rm 1},
    Kaipeng Zhang\textsuperscript{\rm 1}\thanks{Corresponding author},
    Yue Yang\textsuperscript{\rm 1,\rm 2},
    Hao Zhang\textsuperscript{\rm 1, \rm 3},
    Yu Qiao\textsuperscript{\rm 1}
}
\affiliations {
    \textsuperscript{\rm 1}OpenGVLab, Shanghai AI Laboratory\\
    \textsuperscript{\rm 2}Shanghai Jiao Tong University\\
    \textsuperscript{\rm 3}Institute of Artificial Intelligence and Robotics, Xi’an Jiaotong University\\

    gin2pws@gmail.com, 
    zhangkaipeng@pjlab.org.cn, 
    yang-yue@sjtu.edu.cn, 
    zhanghao520@stu.xjtu.edu.cn, 
    qiaoyu@pjlab.org.cn
}

\usepackage{bibentry}

\begin{document}

\maketitle

\begin{abstract}
Vision-language foundation models have been incredibly successful in a wide range of downstream computer vision tasks using adaptation methods. 
However, due to the high cost of obtaining pre-training datasets, pairs with weak image-text correlation in the data exist in large numbers. We call them weak-paired samples. Due to the limitations of these weak-paired samples, the pre-training model are unable to mine all the knowledge from pre-training data. The existing adaptation methods do not consider the missing knowledge, which may lead to crucial task-related knowledge for the downstream tasks being ignored. To address this issue, we propose a new adaptation framework called Data Adaptive Traceback (DAT). Specifically, we utilize a zero-shot-based method to extract the most downstream task-related subset of the pre-training data to enable the downstream tasks. Furthermore, we adopt a pseudo-label-based semi-supervised technique to reuse the pre-training images and a vision-language contrastive learning method to address the confirmation bias issue in semi-supervised learning. We conduct extensive experiments that show our proposed DAT approach meaningfully improves various benchmark datasets performance over traditional adaptation methods by simply.
\end{abstract}

\section{Introduction}

Vision-language ``foundation models" (hereinafter referred to as ``foundation models"), such as CLIP~\cite{CLIP}, ALIGN~\cite{ALIGN}, and Florence~\cite{florence}, have attracted substantial research interest in recent years, demonstrating inspiring outcomes across different computer vision tasks, \eg, image classification~\cite{tipadapter}, object detection~\cite{swintransformer}, and semantic segmentation~\cite{denseclip}. Foundation models are typically pre-trained using large amounts of image-text pairs (usually in millions or billions), such as CC12M~\cite{cc12m}, YFCC 100M~\cite{yfcc100m}, and LAION-400M~\cite{laion400}. The pre-training enables foundation models to obtain high accuracy on downstream tasks using little or even no downstream training data with simple adaptation methods. In this paper, we focus on downstream image classification for convenience.

   

\begin{figure}[ht]
\centering
\begin{subfigure}{\linewidth}
  \centering
  \includegraphics[width=1.01 \linewidth]{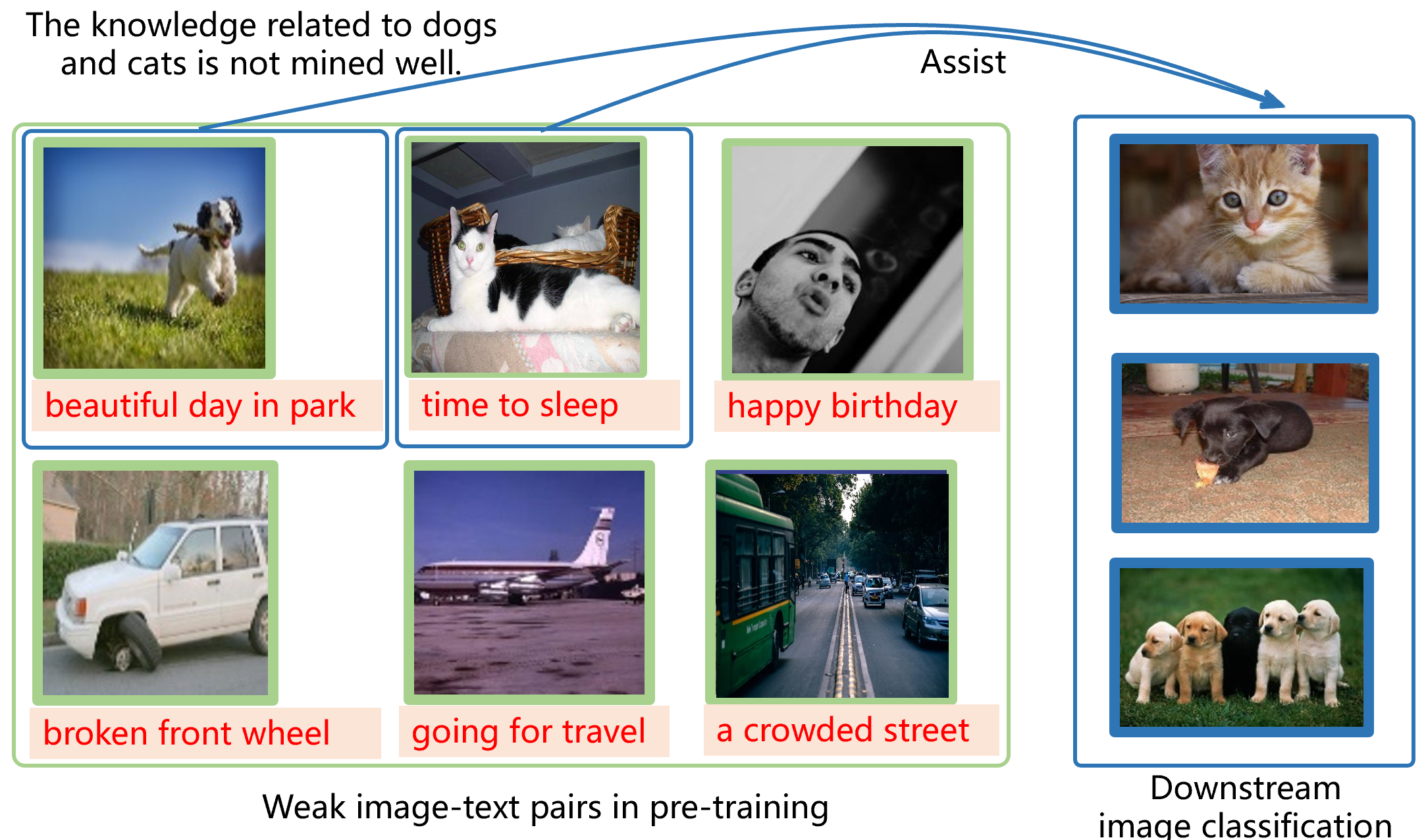}
  \subcaption{}
\end{subfigure}
\hspace{0.5cm} 
\begin{subfigure}{\linewidth}
  \centering
  \includegraphics[width=1.01 \linewidth]{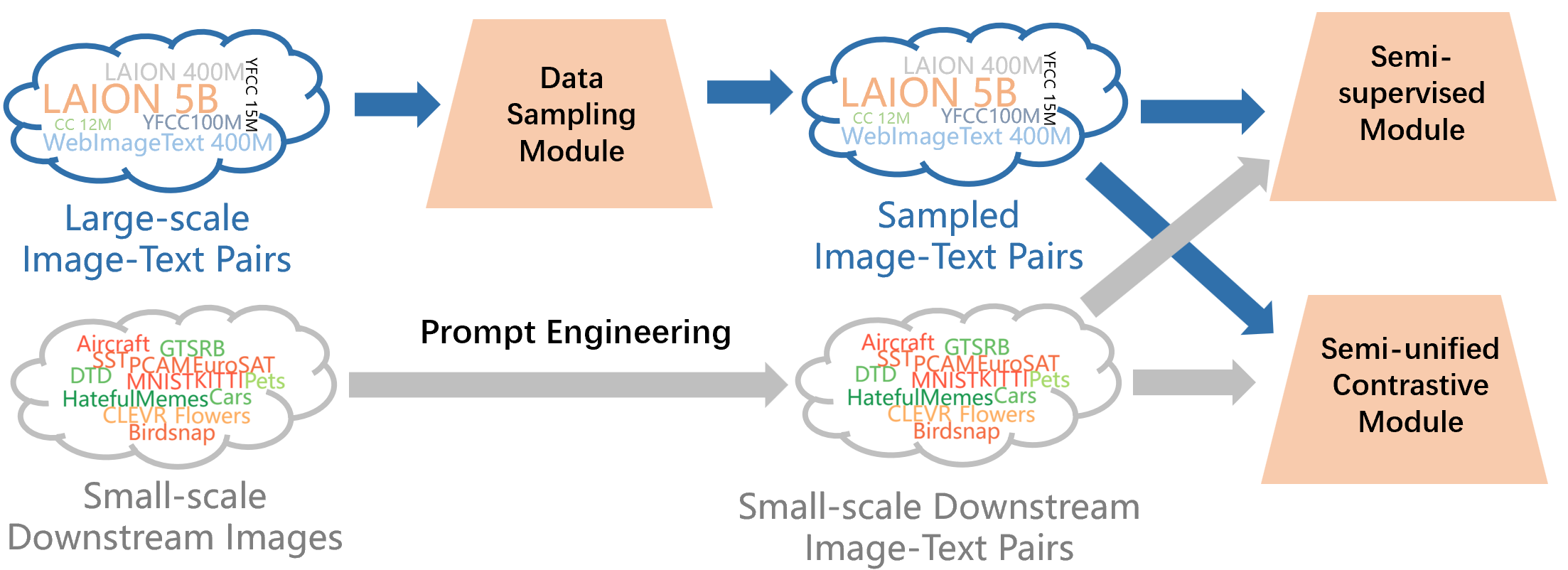}
  \subcaption{}
\end{subfigure}
\caption{(a) An illustration of our motivation. The weak pre-training image descriptions lead the model to ignore some knowledge (e.g., dog and cat in the figure), which may be related to a downstream task. (b) We propose Data Adaptive Traceback (DAT) to retrieve downstream-most-related pre-training data efficiently through the data sampling module and learn them effectively through the semi-supervised module and the semi-unified contrastive module. Please see Figure \ref{fig3}, \ref{fig4}, \ref{fig111} and corresponding sections for more details.}
\label{fff1}
\end{figure}

However, large-scale clean pre-training datasets are difficult and expensive to produce. The pre-training datasets used by many foundation models such as YFCC100M, and LAION400M contain a lot of weak-paired samples. For example, as shown in Figure ~\ref{fff1} (a), the text description is `` what a beautiful day, lets go for a walk in park" and the paired image is a dog playing on the grass. The text description lacks the key information dog and thus the foundation model's attention to the dog is significantly reduced. 
Limited by these weak-paired samples, we believe the model has not mined all the information from the pre-training data during pre-training phase. Therefore, the performance of downstream tasks is limited by directly transferring the knowledge learned by the foundation model.
The existing ``pre-training + adaptation" framework do not consider the incomplete knowledge and we believe it will be an important reason why the model is limited in performance in downstream tasks.  To remedy this shortcoming, we propose to reuse pre-training data in the adaptation process to learn the neglected downstream-related knowledge in the pre-training phase. 

Naturally, semi-supervised learning arises as a solution to reuse pre-training images in the adaptation process. Nevertheless, several points ought to be considered. 
(1) How to reduce the computational cost of reusing the pre-training data during adapatation?
(2) How to overcome the class mismatch and data distribution shift between the pre-training data and the downstream labeled data, and extract useful features for the classification task?
(3) How to leverage the captions and the pre-trained text encoders associated with the pre-trained images to enhance the adaptation process?

To address these challenges, we propose a novel method, Data Adaptive Traceback (DAT), which maximizes the exploitation of large-scale pre-training image-text data during the adaptation stage for foundation models. DAT comprises three modules, namely, a data sampling module, a semi-supervised module, and a semi-unified contrastive module (see Figure~\ref{fff1}, (b)). The data sampling module selects a subset of pre-training data that have a small distribution shift to the downstream data using a zero-shot method based on the class and distribution of downstream datasets. Unlike the conventional way of selecting datasets based on image classes, our method also treats each downstream image as a separate category, thus solving the two major problems: class mismatch and distribution gap.
The semi-supervised module applies two forms of image augmentation on pre-training data: weak augmentation is used to generate one-hot pseudo-labels, which serve as supervision for strong augmentation forms. In this way, we can utilize both pre-training data and downstream data in a conventional supervised learning fashion for downstream image classification. 
The semi-unified contrastive module employs contrastive learning to cluster pre-training data and downstream data based on pre-training text and label-based text correlations during the training process. This approach amplifies the representational differences between out-of-distribution and in-distribution data and enriches the representations of in-distribution data from pre-training data based on text descriptions. It also aligns them with downstream data representations to address the confirmation bias\cite{conformationbais} problem in semi-supervised learning in the real-word setting.

In this study, our contributions can be described as follows:
\begin{itemize} 
    \item We introduce a novel method, Data Adaptive Traceback (DAT), that adapts vision-language foundation models to downstream image classification tasks by efficiently exploiting the pre-training knowledge from large-scale pre-training data. This is the first method that leverages the pre-training dataset for adaptation. 
    \item We devise a zero-shot-based sampling method to select a subset of pre-training data that are highly correlated with downstream data, reducing the computational cost and the data distribution shift of adaptation. We also apply a semi-supervised learning approach to reuse the sampled images and their pseudo-labels for downstream image classification. 
    \item We propose a semi-unified contrastive module that alleviates the confirmation bias in semi-supervised learning by clustering pre-training data and downstream data based on pre-training text and label-based text correlations. This module enhances the feature representations of both pre-training data and downstream data and aligns them in a common space. 
    \item We conduct extensive experiments on eight image classification benchmarks and demonstrate that DAT significantly outperforms the conventional fine-tuning method on all of them. 
\end{itemize}


\section{Related Work}
\subsection{Foundation Model}
Foundation model is a large-scale machine model that can adapt to a variety of downstream tasks by performing self-supervised or semi-supervised training on a large amount of data.
The earliest foundation model is BERT~\cite{bert} which is used in the NLP community. BERT is a bidirectional transformer model which is pre-trained on two tasks, Masked Language Model and Next Sentence Prediction. The success of BERT and GPT have spurred numerous research works in the NLP field, such as GPT series (GPT~\cite{gpt}, GPT2~\cite{gpt2}, and GPT3~\cite{gpt3},GPT4 ~\cite{openai2023gpt4}), Llama2~\cite{llama}, PaLM~\cite{palm}.
In addition to NLP, the foundation model is also beginning to involve images, speech, multimodality, and other fields. For example, OpenAI's CLIP, a foundation model capable of remarkable zero-shot and few-shot performance, demonstrates the potential of foundation models for the fusion of vision and language. Meta proposes Segment Anything \cite{segment} which can be capable of semantic segmentation of arbitrary images. Microsft proposes VALL-E \cite{valle}, which can generate high-quality personalized speech based on text and voice prompts. In this paper we will mainly explore the performance of vision-language foundation model, CLIP, on downstream tasks.

\subsection{Pre-training and Adaptation}
Pre-training and adaptation are two important steps in domain transfer, which is a technique to learn robust and transferable representations for different tasks. Early methods use fully fine-tuning approach to transfer the previous knowledge to downstream tasks. However, with the explosive increase in the number of pre-trained model parameters, the disadvantages of fully fine-tuning requiring fine-tuning of all parameters are gradually magnified, and reseachers begin to explore more efficient adaptation methods. K-adapter~\cite{kadapter} embed the Transformer~\cite{transformer} structure directly into the adapter layer which can effectively transfer the knowledge of pre-trained models. VPT~\cite{vpt} freezes the pre-trained Transformer backbone, and achieves performance comparable to or even exceeds fully fine-tune by introducing a small number of task-specific learnable parameters into the input space during the downstream training process. In this paper, we combine our method with some basic considered adaptation method and we show that by reusing the pre-training data in adaptation process, we can achieve better performance.

\subsection{Contrastive Learning}
Contrastive learning aims to learn a discriminative representation by minimizing the distances between similar data (positive pairs) while maximizing the distances between dissimilar data (negative pairs). Compared with supervised learning with labeled images, recent contrastive learning methods in the image field can train comparable even stronger feature extractors (i.e., pre-trained models) using large-scale unlabeled images. For instance, early works like ~\cite{insdisc} have proposed instance discrimination, where each sample is treated as a separate class and data augmentation is used to generate the positive samples. Moco ~\cite{moco} uses a dynamic queue to store the negative samples and uses the momentum update method to update the parameters of the model. Simclr~\cite{simclr} uses two different image augmentations to the same image in a batch and a newly add projection head to calculate the contrastive loss. Furthermore, contrastive learning has gained attention in the vision-language domain. CLIP and ALIGN use language-image contrastive learning methods to cluster visual and text representations by training on large-scale image-text pairs crawled from the web. Unicl~\cite{unicl} define a triplet image-text-label space to build a unified contrastive learning framework. In our DAT method, we utilize the vision-language contrastive learning method to improve the accuracy of the adaptation process.

\section{Proposed Method} 

\subsection{Data Sampling Module}\label{sec32}
\begin{figure}[t]
\begin{center}
   \includegraphics[width=1.0\linewidth]{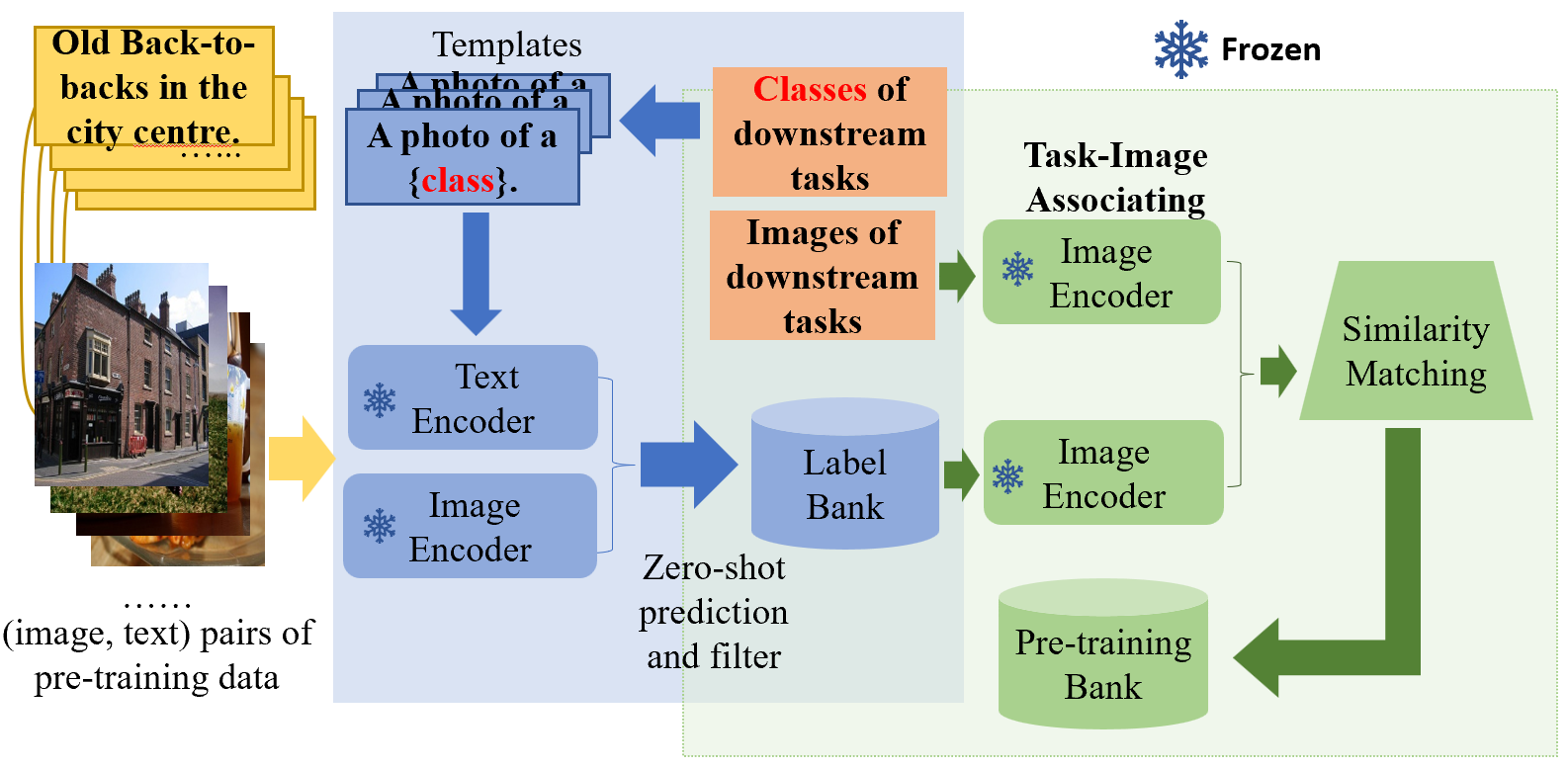}
\end{center}
   \caption{
   The pipeline of the data sampling module. By two consecutive samplings, we sample a subset from large-scale pre-training data, which resembles the distribution and class of downstream images.}
\label{fig3}
\end{figure}

Data sampling module is used to select data with a relatively small class mismatch and distribution shift to downstream data from large amounts of pre-training data. It consists of two steps. The details are shown in Figure~\ref{fig3}. 

Firstly, for class mismatch (pre-training data contains much more classes than downstream data), inspired by CLIP's impressive performance in zero-shot image classification, we perform zero-shot classification to sample data belonging to downstream classes. We introduce the details below. 
In the first step, we rely on the class of downstream data. The same as CLIP's zero-shot test, we use a text template like ``a photo of $\{class\}$" to generate descriptive text for each class. For $C$ categories in the downstream data, we generate $C$ templates, and then we use the pre-trained text encoder of CLIP to get the text features 
$f \in\mathbb{R}^{{C}\times {d}}$, where $d$ is the dimension of text features. 
For pre-training images, we use CLIP's pre-trained image encoder to extract the visual features $v \in\mathbb{R}^{m\times d}$, where $m$ is the total number of pre-training images. Then we calculate the similarity score matrix $S$ of texts and images by $S = {\overline{v}}\cdot {\overline{f}^T}$, where ${\overline{f}}=\frac{f}{\Vert f \Vert}$, ${\overline{v}}=\frac{v}{\Vert v \Vert}$, $S\in\mathbb{R}^{m\times C}$. Then, we take the top k largest values for each column according to the similarity scores. We store the sampled $k \times {C}$ images in the Label Bank. Considering that directly taking the top k maximum values will have some repeated images, we first classify each picture into a certain category by the maximum value of each row. By this step, we sampled $k \times {C}$ different images.

Secondly, for distribution shifts between the pre-training images and the downstream images, in the second step, we treat each downstream image as an independent category and use the same image encoder to get its visual features. After that, we compute the similarity scores between the sampled pre-training images and the downstream images. A high similarity score means more likely to be an in-distribution example. Then, we do a similar sampling process as the first step but it is based on the above similarities.

After the two sampling steps, we get the pre-training bank which stores a subset of the pre-training data. It is downstream data-related and will be used in the next modules. In the following sections, we will present how to use the pre-training bank data during adaptation.

\subsection{Semi-supervised Module}
The semi-supervised module aims to learn downstream task-related knowledge from the pre-training images in a semi-supervised manner. Specifically, we generate downstream task-related pseudo-labels for the sampled pre-training bank, then we train the model in a supervised-learning manner using the pre-training bank and the labeled downstream data. We consider the downstream data with $n$ examples, $\mathcal{D}_{l} = \{(x_{l}^{i},y_{l}^{i}): i\in (1,\ldots, n)\}$, where $x_l^i$ is the $i$-th image in downstream images and $y_l^i$ is its labels. Similarly, let $\mathcal{D}_u = \{(x_u^j,t_u^j): j\in (1,\ldots, m)\}$ be the data from the pre-training bank, where $x_u^j$ is the $j$-th image in the pre-training bank and $t_u^j$ is its corresponding text description and $m$ is the total number of pre-training bank data. 

\begin{figure}[t]
\begin{center}
   \includegraphics[width=1.0\linewidth]{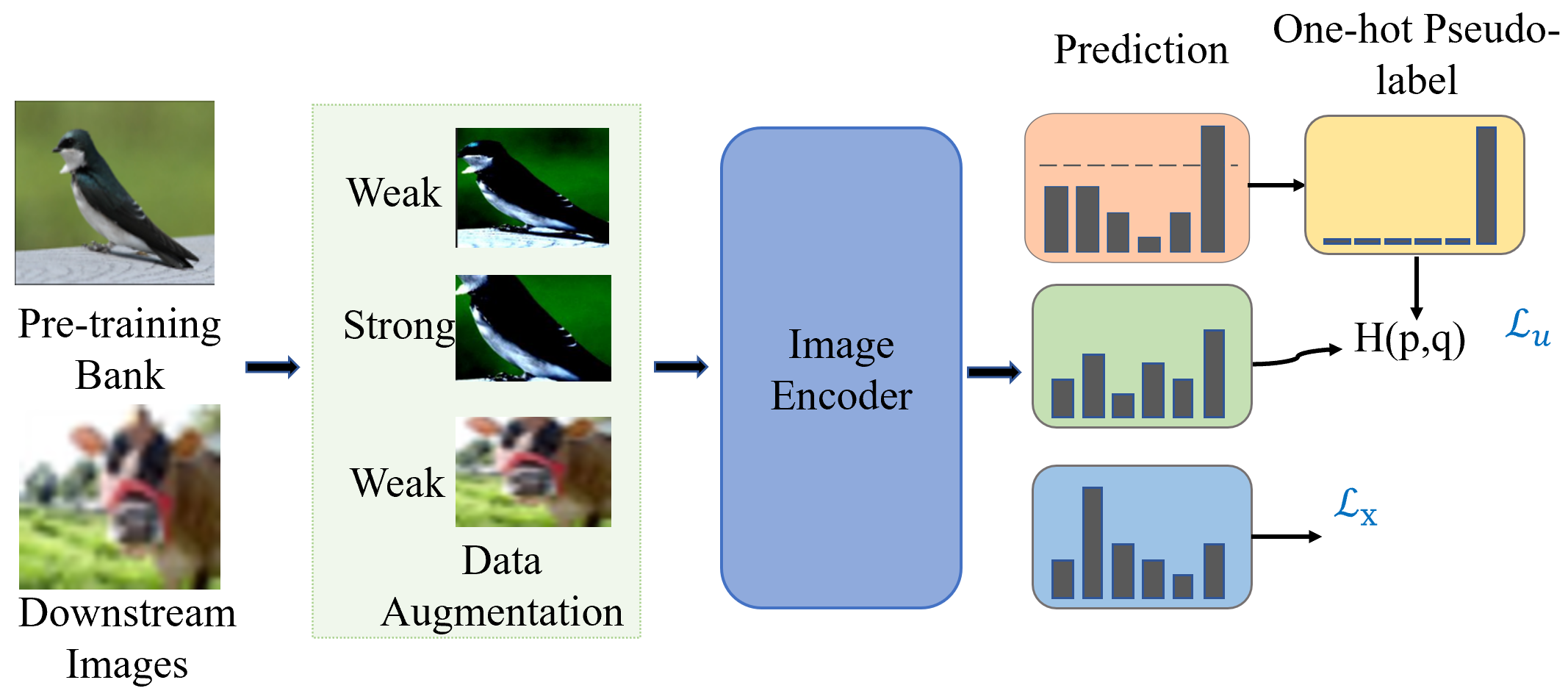}
\end{center}
   \caption{
   The pipeline of the semi-supervised module. With two data augmentation approaches, we can provide pseudo-labels to the pre-training bank images so that the pre-training bank images can be trained in a conventional supervised learning manner. }
\label{fig4}
\end{figure}

As shown in Figure~\ref{fig4}, we follow FixMatch~\cite{fixmatch} to do pseudo-labeling for the pre-training bank data. In our setting, each unlabeled image $x_u^j$ will have two augmentation forms including a weak augmentation $Aug_w(\cdot)$ and a strong augmentation $Aug_s(\cdot)$. For downstream labeled images $x_l^i$, we apply the same weak augmentation $Aug_w(\cdot)$ as unlabeled to get a more noisy form.
The whole module contains a supervised loss $\mathcal L_x$ and an unsupervised loss $\mathcal L_u$. The $\mathcal L_x$ is a standard cross-entropy loss for the labeled examples. Mathematically, for labeled images in a batch $B$, we have 
\begin{equation}
    \mathcal{L}_x = \frac {1}{B}{\sum_{i=1}^{B}}H(y_l^i,P_{cls}(Aug_w(x_l^i))),
\end{equation}
where $P_{cls}(\cdot)$ is a classification head.

$\mathcal L_u$ relies on pseudo label $\hat{q}_j=argmax(p_j)$ from the model's prediction $p_j$ =$P_{cls}(Aug_w(x_u^j))$ on the weak view of image $x_u^j$. A confidence threshold $T_{thresh}$ is used to determine whether this image $x_u^j$ belongs to a class of downstream data. Only unlabeled examples that satisfy $\hat{q}_j$\textgreater$T_{thresh}$ will be considered as in-distribution examples and be retained for consistency training. This pseudo-label will then be used as a supervision to the strong augmentation form $Aug_{s}(x_u^{j})$ and thus we can calculate the classification loss of unlabeled images using simple cross-entropy:
\begin{equation}
   \mathcal L_{u} = \frac {1}{\mu B}{\sum_{j=1}^{\mu B}}\mathbb{I}(\hat{q}_j \ge t )H({\hat{q}_j},P_{cls}(Aug_s(x_u^j))),
\end{equation}
where $\mu$ is the ratio of unlabeled data and labeled data in a batch.

However, current semi-supervised learning methods do not perform well on real-world data, which is due to the inherent problem of confirmation bias in semi-supervised learning. 

\subsection{Semi-unified Contrastive Module}
\begin{figure}[t]
\begin{center}
   \includegraphics[width=1.0\linewidth ]{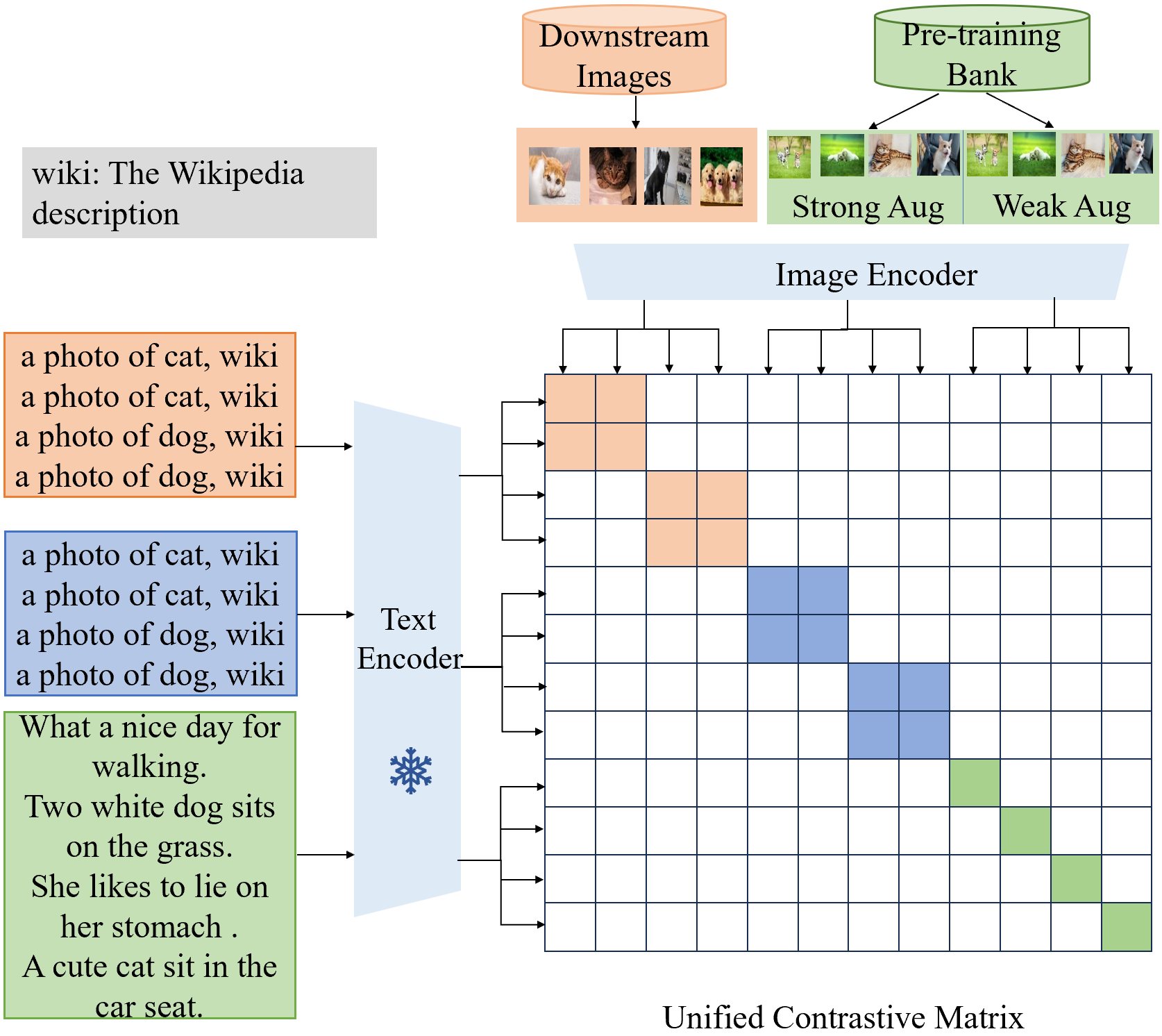}
\end{center}
   \caption{
   Explanation of semi-unified contrastive module. We combine semi-supervised module to map downstream images and pre-training bank datasets into a common image-text-label data space by means of prompt engineering. We cluster the pre-training bank images and down-
stream images by adding image-text contrast loss in the adaptation process to solve the confirmation bias problem.
   }
\label{fig111}
\end{figure}

The semi-unified contrastive module is used to alleviate the influence of conformation bias under the supervision of texts. 
We cluster the pre-training bank images and downstream images by adding image-text contrast loss in the adaptation process to make the in-distribution examples closer and the out-of-distribution examples further. 

As shown in Figure \ref{fff1}, following UniCL~\cite{unicl}, we use a triplet image-text-label data space to unify the pre-training data and downstream data. For the target downstream data, we use prompt engineering to generate the corresponding text descriptions. Specifically, we use the template ``a photo of {cat}" to describe those images which belong to cat. To enrich the semantic information of the text descriptions, we add the definition of the category from Wikipedia after each template. Then our downstream data $\mathcal{D}_l = \{(x_l^i,y_l^i): i\in (1,\ldots, n)\}$ becomes triplet set $\mathcal{D}_l = \{(x_l^i,t_l^i,y_l^i): i\in (1,\ldots, n)\}$, where $t_l^i$ is the generated text description of $x_l^i$ by prompt engineering method. For the source pre-training bank data, unlike UniCL, we combine the semi-supuervised module to generate the triplet data. Specifically, for the weak augmentation view of samples that satisfying $\hat{q}_j$\textgreater$T_{thresh}$, we use the same template engineering method to generate a new text description and thus we create a new triplet $\mathcal{D}_w = \{(x_w,t_w,y_w)\}$, where $x_w$ is the weak view of $x_u$, $t_w$ is the generated text and $y_w$ is the pseudo-label. For the strong augmentation view of all pre-training bank samples,  we generate new labels $y_s^j = j+max(y_l)$ since each image-text pairs are unique. Thus we get the strong format triplet $\mathcal{D}_s = \{(x_s,t_s,y_s)\}$, where $x_s$ is the strong augmentation samples from the pre-training bank, $t_s$ is the original text descriptions.



For both target and source triplets in a batch, each image $x_l(x_w,x_s)$ is encoded by an image encoder $f_\theta$ to get the visual feature $V=f_\theta(x_l)$ and each text description $t_l(t_w,t_s)$ is encoded by a frozen text encoder to get the text feature $T=f_\phi(t_l)$.
Then these features, \ie, $V$ and $T$ are mapped to an unit hyper-sphere with square normalization, ${{v}}=\frac{V}{\Vert V \Vert}$, ${{t}}=\frac{T}{\Vert T \Vert}$.

For all $N$ image-text pairs in a batch($N>B$, becase this $N$ contains the pseudo-labe-based image-text pairs), we can get a feature affinity matrix $S$ by a dot product of visual features $I$=\{$v_i$: $i$=1,...,$N$\} and text features $D$=\{$t_i$: $i$=1,...$N$\} with a temperature constant, as follows:

\begin{equation}
    s_{ij} = exp(\frac{\bm{v}_i\bm{t}_j^T}{\tau})
\end{equation}
where $\tau$ is a temperature coefficient. 
For contrastive loss, it is a bidirectional InfoNCE\cite{infonce} loss that contains two terms:
\begin{equation}
    \mathcal{L}_{con} = \mathcal{L}_{i2t}+\mathcal{L}_{t2i}.
\end{equation}
$\mathcal{L}_{i2t}$ is an image-to-text contrastive loss to align matched images in a batch with given texts. Mathematically,
\begin{equation}
  \mathcal{L}_{i2t}=-\sum_{i\in {N}}\frac{1}{|\mathcal{P}(i)|}\sum_{k\in \mathcal{P}(i)}\log \frac{\exp(\tau \bm{t}_i^T\bm{v}_k)}{\sum_{j\in {N}}\exp(\tau \bm{t}_i^T\bm{v}_j)},
\end{equation}
where $k\in \mathcal{P}(i)=\left \{ k|k\in {N},y_k=y_i\right \}$.

Similar to $\mathcal{L}_{i2t}$, $\mathcal{L}_{t2i}$ is text-to-image contrastive loss to align matched texts to given images.
Mathematically,
\begin{equation}
  \mathcal{L}_{t2i}=-\sum_{j\in {N}}\frac{1}{|\mathcal{P}(j)|}\sum_{k\in \mathcal{P}(j)}\log \frac{\exp(\tau \bm{t}_k^T\bm{v}_j)}{\sum_{j\in {N}}\exp(\tau \bm{t}_k^T\bm{v}_j)},
\end{equation}
where $k\in \mathcal{P}(j)=\left \{ k|k\in {N},y_k=y_j\right \}$.

When the pre-training image $u_s^j$ is the in-distribution example, its corresponding text $ t_s^j$ is similar to the downstream text description $t_l^i$. For those out-of-distribution examples, there is a big gap between the two different sources' texts. 

The total loss can be formulated as follows, respectively:
\begin{equation}
  \mathcal{L}=\mathcal{L}_{x}+\eta\mathcal{L}_{u}+\lambda\mathcal{L}_{con},
\end{equation}
where $\eta$ and $\lambda$ are the weights for semi-supervised loss and contrastive loss.

\begin{table*}[t]
    \centering
    \small
    \resizebox{\textwidth}{!}{
    \begin{tabular}{llllllllllc}
    \hline
        \textbf{Method} & \textbf{CIFAR10} & \textbf{CIFAR100} & \textbf{FGVC Aircraft} & \textbf{Oxfordpets} & \textbf{Stanford Cars} & \textbf{DTD} &
        \textbf{Food101} &
        \textbf{SUN397} & \textbf{Avg}      
        \\ 
        \hline
        Fine-tuning & 95.52 &79.50 & 78.37 & 85.53 & 82.08 & 70.59 & 86.90  & 69.26&80.97\\
        Fixmatch+RS &95.69& 80.33 & 78.22 & 86.13 & 80.31 & 72.43&86.55 & 69.41&81.13\\
        CCSSL+RS& 95.22&78.34&78.31&86.29&80.33&71.38&86.71&69.38&80.75\\
        Fixmatch+DS &95.82& 80.15 & 78.19 & 86.37 & 82.35 & 72.43&86.59 & 69.03&81.37\\
        CCSSL+DS& 95.50&80.32&78.52&86.95&82.64&72.56&87.21&68.35&81.51\\
        DAT &  \textbf{96.21}&\textbf{81.13}& \textbf{78.85} & \textbf{87.98} & \textbf{83.75} & \textbf{73.88} & \textbf{87.90} & \textbf{70.17}&\textbf{82.48}\\
    \hline
    \end{tabular}
    }
    \caption{Evaluation on eight image classification benchmarks. All methods use the ResNet-50 image encoder of OpenCLIP pre-trained on YFCC15M. For the semi-supervised learning method Fixmatch~\cite{fixmatch} and CCSSL~\cite{yang2022class}, we randomly selected(RS) the same amount of pre-training data as our Data Sampling Module(DS).}
    \label{tab2}
    
\end{table*}

\begin{table}[t]

\centering
\small
\setlength{\tabcolsep}{3mm}\begin{tabular}{ccc}
\hline
Method& Oxford Pets & DTD \\
\hline
Fine-tuning Baseline&85.53&70.59\\
\hline
Random Sampling&86.32&73.14\\
Data Sampling Moudle&\textbf{87.98}&\textbf{73.88}\\
\hline
\end{tabular}
\caption{Ablation experiment results for the Data Sampling Module on Oxford Pets and DTD.}
\label{tab3}
\end{table}

\begin{table}[t]
\centering
\small
\setlength{\tabcolsep}{3mm}\begin{tabular}{ccccc} 
\hline   
S-loss & U-loss & C-loss & Oxford Pets  & DTD\\
\hline
\checkmark &  &  & 85.53 & 70.59\\
\checkmark&\checkmark&&86.37&72.43\\
\checkmark&&\checkmark&86.75&73.40\\
\checkmark&\checkmark&\checkmark&\textbf{87.98}&\textbf{73.88}\\
\hline 
\end{tabular}
\caption{Ablation experiments on loss functions. S-loss, U-loss, and C-loss represent supervised loss, unlabeled loss, and contrastive loss respectively.}
\label{tab5}
\end{table}

\begin{table}[t]
\begin{center}
\small
\centering
\setlength{\tabcolsep}{3mm}\begin{tabular}{crrrr}
\hline
Ratio & \multicolumn{2}{c}{Oxford Pets} & \multicolumn{2}{c}{DTD} \\ 
($\mu$) &   \multicolumn{1}{c}{0.6} & \multicolumn{1}{c}{0.95}  & \multicolumn{1}{c}{0.6} & \multicolumn{1}{c}{0.95} \\
\hline 
2 &86.07&87.19&72.61&73.46 \\
{3} &\textbf {86.32}&\textbf{87.98}&\textbf{73.83}&72.93\\
{4} &85.64&86.86&72.61&\textbf{73.88}\\
{5} &85.75&86.89&72.07&73.03\\
{6} &85.88&87.19&72.87&73.35\\
{7} &85.47&87.00&72.93&73.62\\
\hline
\end{tabular}
\end{center}
\caption{Experiment on different ratios of unlabeled data on DTD and Oxford Pets with different thresholds.}
\label{tab6}
\end{table}

\begin{table}[t]
\centering
\small
\setlength{\tabcolsep}{3mm}\begin{tabular}{ccc}
\hline
$T_{thresh}$& Oxford Pets & DTD \\
\hline
0.50&86.51&72.93\\
0.60&85.64&73.24\\
0.70&86.43&73.35\\
0.80&86.02&72.93\\
0.90&\textbf{86.86}&73.46\\
0.95&86.67&\textbf{73.88}\\
\hline
\end{tabular}
\caption{Experiment on different thresholds. We test the optimal threshold ranging from 0.5 to 0.95.}
\label{tab7}
\end{table}

\section{Experiment}
\begin{table*}[t]
\centering
\small
    \resizebox{\textwidth}{!}{
    \begin{tabular}{llllllllllc}
    \hline
        \textbf{Method} & \textbf{CIFAR10} & \textbf{CIFAR100} & \textbf{FGVC Aircraft} & \textbf{Oxfordpets} & \textbf{Stanford Cars} & \textbf{DTD} &
        \textbf{Food101} &
        \textbf{SUN397} & \textbf{Avg}        
        \\ 
        \hline
        linear-probing &92.15&75.70&50.68&87.19& 87.54& 76.81&85.76&74.47&78.79\\
        Fine-tuning & 96.93 &86.08 & 71.86 & 87.35 & 87.65 & 72.34 & 85.78  & 69.41&82.18\\
        bias-tuning & 96.68&85.20&65.41&88.58&89.70&79.31&87.01&74.02&83.24\\
        adapter-tuning & 96.90&85.32&70.69&88.83&89.75&77.45&87.83&74.16&83.87\\
        DAT-finetuning&\textbf{98.22}&\textbf{88.08}&\textbf{77.59}&\textbf{89.86}&\textbf{91.31}&\textbf{79.01}&\textbf{87.62}&\textbf{74.29}&\textbf{85.75}\\
        DAT-bias tuning &\textbf{97.21} &\textbf{86.78}&\textbf{68.71}&\textbf{89.42}&\textbf{89.38}&\textbf{80.56}&\textbf{87.76}&\textbf{75.23}&\textbf{84.31}\\
        DAT-adapter&\textbf{97.85}&\textbf{86.29}&68.65&\textbf{89.92}&\textbf{90.34}&\textbf{79.75}&\textbf{88.12}&\textbf{75.10}&\textbf{84.50}\\
    \hline
    \end{tabular}
    }
    \caption{The ViT-B-32 model test results. We combine our DAT with three commonly used adaptation methods: fine-tuning, bias-tuning, and adapter-tuning. }
    \label{tab7-1}

\end{table*}

\begin{table*}[t]
\centering
\small
    \resizebox{\textwidth}{!}{
    \begin{tabular}{llllllllllc}
    \hline
        \textbf{Method} & \textbf{CIFAR10} & \textbf{CIFAR100} & \textbf{FGVC Aircraft} & \textbf{Oxfordpets} & \textbf{Stanford Cars} & \textbf{DTD} &
        \textbf{Food101} &
        \textbf{SUN397} & \textbf{Avg}        
        \\ 
        \hline
        Fine-tuning & 96.95 &85.23 & 73.03 & 91.22 & 85.77 & 72.50 & 87.35  & 69.10&82.64\\
        DAT-finetuning&\textbf{98.15}&\textbf{87.96}&\textbf{75.70}&\textbf{93.31}&\textbf{89.24}&\textbf{77.34}&\textbf{88.92}&\textbf{74.38}&\textbf{85.63}\\
    \hline
    \end{tabular}
    }
    \caption{Results of reusing other pre-training datasets. We use the ViT-B-32 model pre-trained on WebImageText data as our backbone and reuse the LAION400M data.}
    \label{tabpws}
\end{table*}

In this section, we will first introduce our experimental setup. Then we show our results on eight image classification benchmarks using RN50 pre-trained on YFCC15M\cite{yfcc100m} data and ViT pre-trained on LAION-400M. We present ablation and hyperparameters experiments on our proposed three modules. 

\subsection{Experimental Setup}
In our study, we use CLIP as the foundation model to evaluate our adaptation method. To evaluate our adaptation method, we use CLIP as the foundation model. The original pre-training data for CLIP is WebImageText-400M~\cite{CLIP}, which contains 400 million image-text pairs. Given the size of this dataset, we opt to follow OpenClip~\cite{openclip} and use a ResNet-50~\cite{resnet} image encoder pre-trained on YFCC15M~\cite{yfcc100m} in our experiments. We evaluate our method on eight widely-used image classification benchmarks, including CIFAR10~\cite{cifar}, CIFAR100~\cite{cifar}, FGVA-aircraft~\cite{aircraft}, Oxford Pets~\cite{oxfordpets}, Stanford Cars~\cite{stanfordcars}, DTD~\cite{dtd}, Food101~\cite{food}, and SUN397~\cite{sun397}. During training, we set $\lambda = 1$ with $T_{thresh}=0.95$ due to the high noise level of the pre-training data and we also set $\eta =1 $. In our experiment, we take fine-tuning as an example to show our test results. We train 12 epochs because of the fast convergence speed of the pre-trained model.

\subsubsection{Baseline}

Fine-tuning is the most widely-used adaptation method, and we incorporate our method with CLIP and conventional fine-tuning. Firstly, we use OpenCLIP's ResNet-50 image encoder pre-trained on YFCC15M as backbone and fine-tune the parameters using downstream labeled images. In addition to the conventional fine-tuning method, we compare our results with commonly used semi-supervised methods Fixmatch~\cite{fixmatch} and CCSSL~\cite{yang2022class}. We chose these two methods for several considerations. Fixmatch is a classic and effective semi-supervised learning method that has been utilized in many semi-supervised learning papers. We choose CCSSL because it can be used with any pseudo-label-based semi-supervised learning method to alleviate confirmation bias in the training process. However, we state that our DAT framework can be combined with not only these two semi-supervised learning methods, it can be combined with any pseudo-label based semi-supervised learning method.

\subsubsection{Data Sampling Module}
We use the data sampling module to select datastream-related samples from pre-training data. As introduced in Section~\ref{sec32}, this module has two steps. In the first step, we ensure that each text feature corresponds to one class by averaging multiple text representations for each class. We  sample eight times more data than the downstream data in this step. In the second step, we further select half of the images sampled in the first step, leading to a final sample size that is four times larger than the downstream data.

\subsection{Evaluation on Image Classification Benchmarks}

Table~\ref{tab2} summarizes our evaluation results on image classification. Our DAT method outperformed fine-tuning and semi-supervised learning methods on all benchmarks. Specifically, on datasets with relatively low prediction accuracy such as CIFAR100, FGVC-Aircraft, Stanford Cars, DTD, and SUN397, DAT achieved increases of $+1.63\%$, $+0.48\%$, $+1.67\%$, $+3.29\%$, and $+0.91\%$, respectively, compared to fine-tuning. On other datasets, namely CIFAR10, Oxford Pets, and Food101, DAT achieved improvements of $+0.69\%$, $+2.45\%$, and $+0.51\%$, respectively, over the fine-tuning method. These results indicate that reusing pre-training data during adaptation can lead to better downstream classification performance. Furthermore, we compare DAT with semi-supervised learning methods, CCSSL and Fixmatch. To make a fairer comparison, for the unlabeled pre-trained data, we used random sampling and DAT sampling methods, respectively. Table ~\ref{tab2} shows that our DAT can achieve better results, which suggests that our DAT can effectively better utilize the pre-trained bank data.



\subsection{Ablation and Hyper-parameters}
\subsubsection{Data Sampling Moudle}

In this section, we evaluate the effectiveness of the data sampling module on two datasets, Oxford Pets, and DTD. Table~\ref{tab3} shows that combining the semi-supervised module and the semi-unified contrastive module with both random sampling and the proposed sampling module can improve the baseline accuracy. Furthermore, the proposed sampling module can select samples that are more relevant to downstream data and can result in further improvements.

\subsubsection{Loss Function}
In this section, we evaluate the effectiveness of the unlabeled loss in the semi-supervised module and the contrastive loss generated in the semi-unified contrastive module. We conduct the experiments on two datasets, Oxford Pets and DTD, and divide them into three aspects.

Table~\ref{tab5} shows that both loss functions are effective. On Oxford Pets, the unlabeled loss and the contrastive loss achieve $+0.84\%$ and $+1.22\%$ improvements, respectively, over the baseline using supervised loss. Using both loss functions achieves $+2.45\%$ improvement. Notably, on DTD, a more challenging dataset with a larger distribution shift, we achieve greater improvements, $+3.29\%$.

\subsubsection{Ratio of Unlabeled Data}

The ratio $\mu$ represents the proportion of sampled pre-trained data and downstream data in a batch during the training process. We conduct ablation experiments on two different thresholds, 0.6 and 0.95, and observe the effects on Table~\ref{tab6}.

Our observations show that a smaller ratio results in the best performance. On DTD, DAT achieves the best performance with $\mu=3$ when $T_{thresh}=0.6$. For a higher threshold of $T_{thresh}=0.95$, we achieve the best performance with $\mu=4$. On Oxford Pets, the results show that our method benefits from a lower ratio, with the best performance occurring when $\mu=3$ for both $T_{thresh}=0.6$ and $T_{thresh}=0.95$. This suggests that a larger proportion of unlabeled data may bring more noise data.

\subsubsection{Threshold for unlabeled loss}

We set the threshold $T_{thresh}$ to range from 0.5 to 0.95 and fix the ratio $\mu=4$ to test the robustness of our method. 

In Table~\ref{tab7}, we find that our DAT performs very stable on two different datasets, Oxford Pets and DTD, with different thresholds $T_{thresh}$. However, the optimal threshold for datasets with different distribution shifts is not consistent.

On Oxford Pets, $T_{thresh}=0.9$ achieves the best performance while $T_{thresh}=0.95$ on DTD. On both datasets, we observe that higher thresholds are more favorable for our DAT.  The results show that downstream data with closer distribution to the pre-trained data needs a smaller $T_{thresh}$, which can prove that our DAT can reduce the harmful influence of noisy data under the supervision of texts.

\section{Larger Data Scaling Results}

In this section, we evaluate our DAT method using larger pre-training datasets, specifically the LAION-400M dataset. We use ViT-B-32 model pre-trained on LAION data as our image encoder and test on above datasets using three adaptation methods: full fine-tuning, bias-tuning\cite{bitfit}, and adapter-tuning\cite{adapter}. Considering computing resources, we utilized a subset of LAION-400M which contains 30 million image-text pairs.
The results are reported in Table~ \ref{tab7-1}.

Table~\ref{tab7-1} demonstrates that our DAT approach significantly improves the performance on all datasets with the three adaptation methods, especially with full fine-tuning. The test results on the eight datasets show an average improvement of +3.57\% when compared to baseline fine-tuning. By comparing the results with (RN50, YFCC15M), the average accuracy (ViT-B-32, LAION-400M subset) increases from +1.45\% to +3.57\%. Hence, our DAT gains are more pronounced with larger networks and reusable data.

Furthermore, we notice that combining DAT with bias-tuning or adapter-tuning does not result in significantly more improvements than either bias-tuning or adapter-tuning alone. We explain this by stating that these two adaptation methods transfer knowledge related to downstream tasks while keeping model parameters unchanged, which limits their capacity to learn new knowledge. Consequently, our experimental results support our initial hypothesis that some knowledge related to downstream tasks may be discarded during the pre-training stage.

\section{Other Pre-training Data }
In this section, we test with other different pre-training datasets to demonstrate that our method can leverage arbitrary image-text pair datasets to improve the performance of downstream classification tasks. 
Specifically, we use the ViT-B-32 pre-trained on the WebImageText-400M dataset as our backbone. Unlike the above method that reuses the WebImageText-400M dataset, we use the LAION400M dataset as our external pre-training dataset and reuse them in adaptation process. The results are shown in Tabel \ref{tabpws}.
From the results, we can observe that our DAT can achieve an average increase of $2.99\%$, which prove that our method can not only learn knowledge from the pre-training dataset, it can be extended to any external image-text dataset.

\section{Limitations and Conclusion}

Our experimental results are subject to potential fluctuations due to the low quality of the pseudo-labels obtained by the thresholding method. Additionally, the text data in the pre-training set contains a significant amount of noise, which can make the training process difficult and lead to convergence issues.

In conclusion, to solve the problem of weak-paired image-text samples from the pre-training dataset, which will result in the foundation model not being able to learn all the knowledge in the pre-training phase.
We propose a novel adaptation method Data Adapative Traceback which can effectively leverage the pre-training data in the adapation phase. By reusing the pre-training image-text pairs, DAT offers a new approach to adaptation that improves upon the most commonly used methods, fine-tuning, bias-tuning, and adapter-tuning. We have demonstrated the effectiveness of DAT on 8 commonly used datasets and established that our approach can facilitate the adaptation process. Our DAT can also be adapted to arbitrary pseudo-labeling-based semi-supervised learning methods, and effectively solves the problem of confirmation bais in semi-supervised learning.

\section{Acknowledgements}
This work is supported in part by the National Key R\&D Program of China (NO. 2022ZD0160100)

\bibliography{pws.bib}

\clearpage

\end{document}